\documentclass[10pt,twocolumn,letterpaper]{article}

\usepackage{wacv}
\usepackage{times}
\usepackage{epsfig}
\usepackage{graphicx}
\usepackage{amsmath}
\usepackage{amssymb}

\wacvfinalcopy %
\ifwacvfinal
\def\assignedStartPage{9876} %
\fi

\ifwacvfinal
\usepackage[breaklinks=true,bookmarks=false]{hyperref}
\else
\usepackage[pagebackref=true,breaklinks=true,colorlinks,bookmarks=false]{hyperref}
\fi

\ifwacvfinal
\else
\fi

\author{Soumya Tripathy\\
Tampere University\\
Finland\\
{\tt\small soumya.tripathy@tuni.fi}
\and
Juho Kannala\\
Aalto University\\
Finland\\
{\tt\small juho.kannala@aalto.fi}
\and
Esa Rahtu\\
Tampere University\\
Finland\\
{\tt\small esa.rahtu@tuni.fi}
}

\date{\today}
\title{FACEGAN: Facial Attribute Controllable rEenactment GAN}
\hypersetup{
 pdfauthor={Soumya Tripathy},
 pdftitle={FACEGAN: Facial Attribute Controllable rEenactment GAN},
 pdfkeywords={},
 pdfsubject={},
 pdfcreator={Emacs 27.1 (Org mode 9.3)},
 pdflang={English}}
\begin{document}

\maketitle

\begin{abstract}
The face reenactment is a popular facial animation method where the person's identity is taken from the source image and the facial motion from the driving image. Recent works have demonstrated high quality results by combining the facial landmark based motion representations with the generative adversarial networks. These models perform best if the source and driving images depict the same person or if the facial structures are otherwise very similar. However, if the identity differs, the driving facial structures leak to the output distorting the reenactment result. We propose a novel Facial Attribute Controllable rEenactment GAN (FACEGAN), which transfers the facial motion from the driving face via the Action Unit (AU) representation. Unlike facial landmarks, the AUs are independent of the facial structure preventing the identity leak. Moreover, AUs provide a human interpretable way to control the reenactment. FACEGAN processes background and face regions separately for optimized output quality. The extensive quantitative and qualitative comparisons show a clear improvement over the state-of-the-art in a single source reenactment task. The results are best illustrated in the reenactment video provided in the supplementary material. The source code will be made available upon publication of the paper.
\end{abstract}

\begin{figure}[!tp]
\centering
\includegraphics[width=1.0\linewidth]{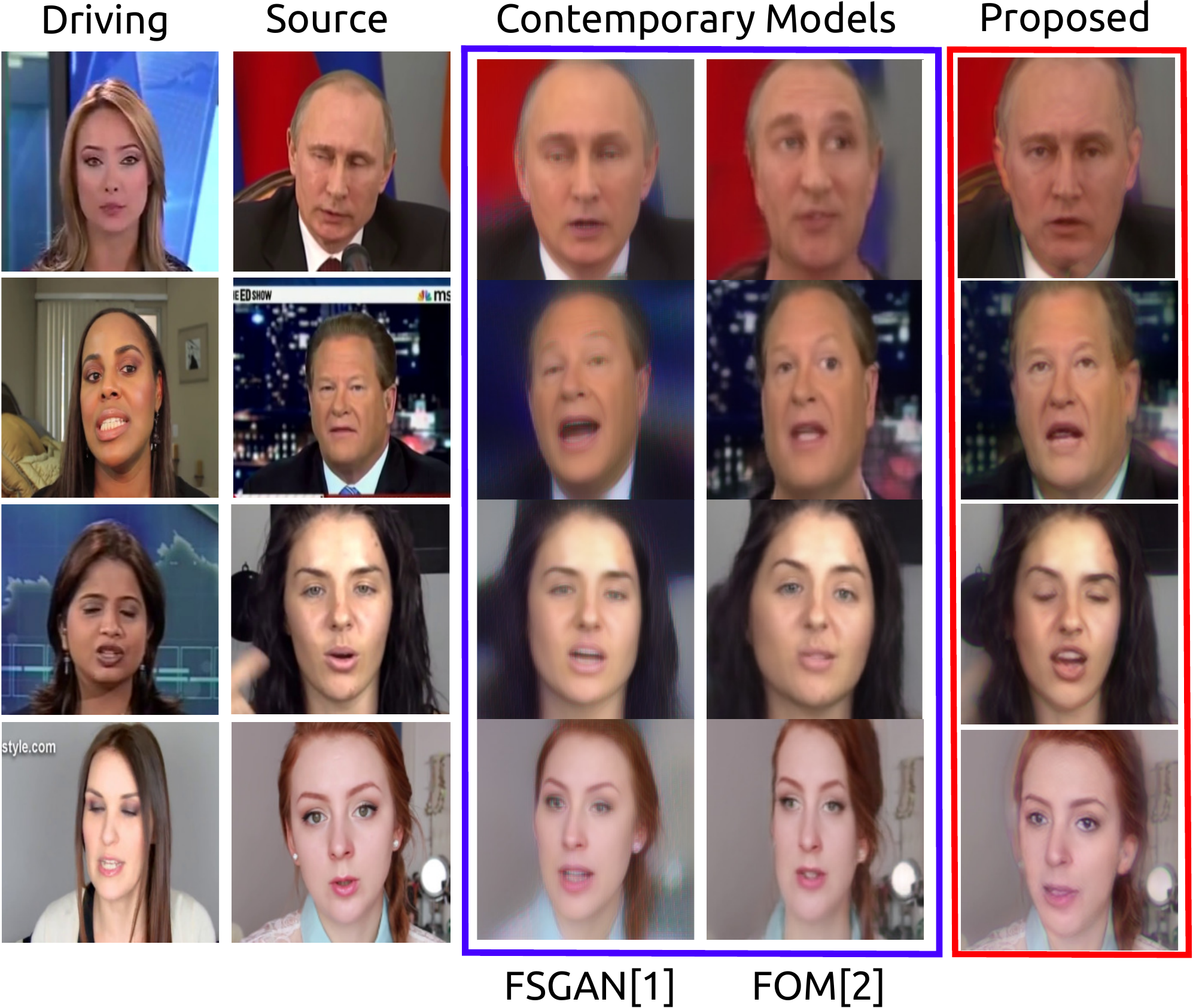}
\caption{\label{fig:org1574033} Examples of the face-reenactment results using the proposed FACEGAN and recent baseline works FSGAN \cite{nirkinFSGANSubjectAgnostic2019} and First-Order-motion-Model (FOM) \cite{siarohinFirstOrderMotion2019}. FACEGAN is able to preserve the source identity faithfully while imposing the driving pose and expression. The baselines suffer particularly in preserving the facial shape of the source.}
\vspace{-0.5cm}
\end{figure}

\vspace{-0.5cm}
\section{Introduction}
\label{sec:orgc3316ef}

Face-reenactment is a process of animating a source face according to the motion (pose and expression) of a driving face. In general, the process involves three major steps: 1) creating a representation of the source face identity, 2) extracting and encoding the motion of the driving face, and 3) combining the identity and motion representations to produce a modified source face. Each part has a significant impact on the output quality.

Number of algorithms, including traditional \emph{3D} face models \cite{thiesFace2FaceRealtimeFace,alexanderDigitalEmilyProject2009}, data driven Neural-Networks \cite{zakharovFewShotAdversarialLearning2019,haMarioNETteFewshotFace2019,pumarolaGANimationAnatomicallyawareFacial2018,nirkinFSGANSubjectAgnostic2019,nirkinFaceSegmentationFace2017,choiStarGANUnifiedGenerative2018,tripathyICfaceInterpretableControllable2020,wilesX2FaceNetworkControlling2018,wuReenactGANLearningReenact2018}, and their combinations \cite{yaoMeshGuidedOneshot2020} have been presented for creating photorealistic face animations. In \emph{3D} face-model based approach \cite{thiesFace2FaceRealtimeFace} , the identity and motion features are encoded with \emph{3D} model parameters. The reenacted face is then rendered using the identity parameters of the source and motion parameters of the driving face. Although this approach results in high quality outputs, they require substantial efforts in obtaining the faithful \emph{3D} representations of the faces. Therefore, such approaches are often limited to a few source identities.

In recent years, the data driven deep neural networks have gained popularity in generating and manipulating images. In particular, the
models using adversarial loss functions \cite{goodfellowGenerativeAdversarialNetworks2014} have obtained highly realistic image synthesis results \cite{karrasStyleBasedGeneratorArchitecture2019,karrasProgressiveGrowingGANs2018}. Similar models are also applied for the face reenactment as demonstrated in \cite{zakharovFewShotAdversarialLearning2019,nirkinFSGANSubjectAgnostic2019,wuReenactGANLearningReenact2018,wilesX2FaceNetworkControlling2018,haMarioNETteFewshotFace2019,zhangFReeNetMultiIdentityFace2020,tripathyICfaceInterpretableControllable2020,siarohinFirstOrderMotion2019,siarohinAnimatingArbitraryObjects2019}. In most of these works, the first step is to represent the source and driving faces with deep networks. These representations can be either latent codes or human interpretable representations like facial landmarks. Subsequently, the facial feature representations are combined by a generator network to produce the final output. Although these
models generalise well to multiple identities, they have numerous challenges discussed in the following.

\begin{figure}[t]
\centering
\includegraphics[width=0.95\linewidth]{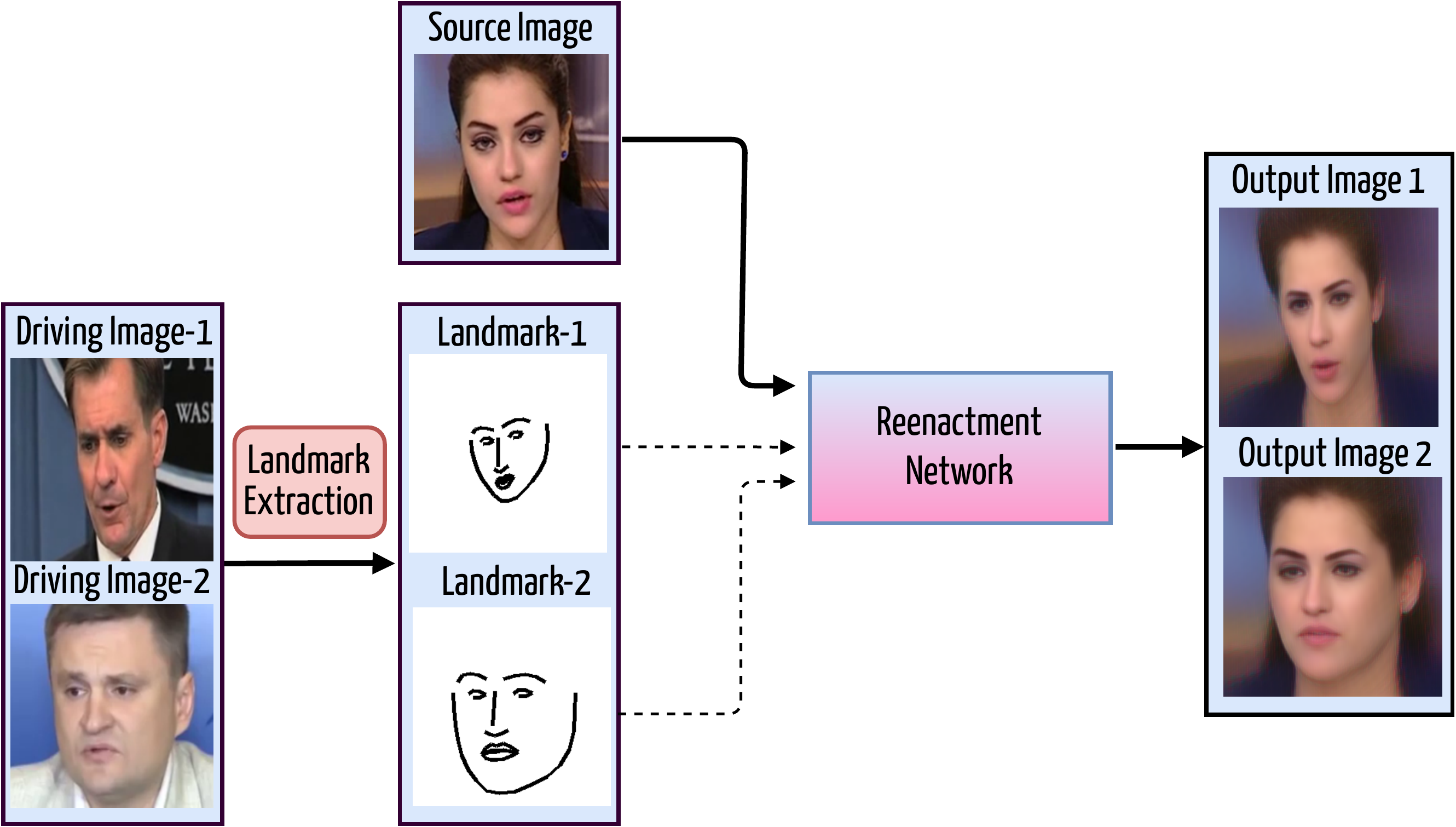}
\caption{\label{fig:org6d9d079} An illustration of the facial structure leaking problem in the landmark based reenactment approach. The facial structure of the generated output clearly follows the shape of the driving face instead of source face as intended.}
\vspace{-0.5cm}
\end{figure}

\vspace{-0.5cm}
\paragraph{Landmark based motion representation}
\label{sec:org78a9baa}

Most recent works \cite{zakharovFewShotAdversarialLearning2019,nirkinFSGANSubjectAgnostic2019} use facial landmark points as the representation of the facial motion (pose and expression). Although landmarks can provide strong supervision of the driving motion, they also contain the overall structure of the driving face in the form of facial contours. If the driving landmarks are directly used to animate the source face, the difference between the facial contours may lead to dramatic distortions in the output result.
Therefore, the gap in the facial shapes usually leads to either losing the source identity or low photorealism. Figure \ref{fig:org6d9d079}
illustrates a practical example of such distortions. For this reason, many works search for driving and source pairs with highly similar face structures. However, this greatly limits the applicability of such methods.
\vspace{-0.7cm}
\paragraph{Learning based representation}
\label{sec:org768e360}
Instead of landmarks, some works \cite{wilesX2FaceNetworkControlling2018,zakharovFewShotAdversarialLearning2019,siarohinAnimatingArbitraryObjects2019} use unsupervisedly learned latent features to represent the facial identity and motion. These representations are often not disentangled with respect to identity, structure and motion. Hence, the driving motion features may contain the driving identity information, which ends up to the final reenacted output. This identity leakage problem is illustrated in figures \ref{fig:rcomp},\ref{fig:org1574033} in the case of X2face \cite{wilesX2FaceNetworkControlling2018} and FOM \cite{siarohinFirstOrderMotion2019}.
\vspace{-0.5cm}
\paragraph{Generalization and inference time training}
\label{sec:org46ba32e}
In some works, the entire reenactment model is trained for a particular source identity \cite{wuReenactGANLearningReenact2018} or the model is fine tuned using a few shot approach at the inference time \cite{zakharovFewShotAdversarialLearning2019}. Although such a strategy
may improve the output quality, it requires resources and time to adapt to a new identity.

\vspace{-0.5cm}
\paragraph{Selective editing}
\label{sec:org02ad039}

It is often desirable to selectively edit the reenactment result, for example, add an extra smile or open the eyes. This type of editing would be easier by directly manipulating the motion representation instead of searching or producing an alternative driving image. To this end, the representation of the driving motion should be interpretable for the human editor. Although landmark points can be easily moved for editing, one needs sophisticated tools to preserve the structure of the source face while adjusting the landmarks. An alternative approach based on Action Units (AUs) is presented in \cite{tripathyICfaceInterpretableControllable2020}, which allows editing without distorting the identity. However, the model was not able to produce high fidelity reenactment results.

In this paper, we propose a Facial Attribute Controllable rEenactment GAN (FACEGAN), which produces high quality source reenactment from various driving pairs even with significant facial structure differences between them. Our model manipulates the source face-landmarks with driving facial attributes to generate a new set of landmarks representing the desired motion with the source identity and structure. This mitigates the identity leakage problem present in many recent works. Furthermore, by representing the motion cues using the action units, we provide a selective editing interface to the reenactment process. The proposed method combines the benefits of the action units (subject agnostic and interpretable) and facial landmarks (strong supervision for image generation) into a single reenactment system. To the best of our knowledge, this is the first work that proposes such a combination for the face reenactment. In addition, FACEGAN decouples and handles facial region and background in two separate branches that help in generating the source background realistically in the final output unlike other models in the literature. Finally, we provide a detailed comparison of our method against the recent state-of-the-art works in a single source image reenactment. The proposed FACEGAN model results in superior performance both in quantitative and qualitative measures.

\section{Related work}
\label{sec:orgd016b53}

Face-reenactment has been actively studied over the years, and several different approaches have been presented. The proposed methods can be roughly grouped into the following three categories.

\vspace{-0.5cm}
\paragraph{Morphable 3-D face models}
Parametric 3D face models are popular tools for face animation and several works \cite{thiesFace2FaceRealtimeFace,bouazizOnlineModelingRealtime2013} have adopted them to face reenactment. These methods start by fitting a 3D morphable face model to source and driving faces. Afterward, the motion and pose parameters of the driving model are transferred to the source model, and the output face with the source identity and the driving motion is rendered. The motion parameters can also be controlled using other modalities such as audio \cite{karrasAudiodrivenFacialAnimation2017}. Although these methods can provide high output quality, they are not easily scaled to a large number of identities.

\vspace{-0.5cm}
\paragraph{Deep generative models}
Variational Auto Encoder (VAE) \cite{kingmaAutoEncodingVariationalBayes2014} is a popular model for image generation and manipulation. VAE encodes the input image into a latent representation vector that is subsequently decoded back to the original input image. The VAE models can be utilised for face animation, by manipulating the latent representation of the source face according to the representation of the driving face \cite{houDeepFeatureConsistent2016}. If the latent representation is disentangled with respect to the identity and motion, the facial movements can be controlled by manipulating the corresponding vector elements of the encoded source face \cite{higginsVVAELEARNINGBASIC2017}. However, the disentangled representation is very difficult to obtain, which often leads to compromises in the output quality.

An alternative encoder-decoder based approach was presented in \cite{wilesX2FaceNetworkControlling2018}. In this work, an encoder-decoder model is applied to warp the input face into an embedding image. Another encoder-decoder network converts the driving image into a warping field that maps the embedding image to the output face. This approach provides better quality compared to the VAE based models, but the driving identity can still easily leak to the reenacted output. This phenomenon is further emphasized if the source and driving faces have large structural or pose differences.

Recently, another warping based reenactment approach was proposed in \cite{siarohinFirstOrderMotion2019}. The method learns a warping field using a set of keypoints from the source and driving images. This approach obtains higher output quality compared to \cite{wilesX2FaceNetworkControlling2018}, but it suffers from similar identity leakage problems. To this end, the authors of \cite{siarohinFirstOrderMotion2019} proposed an alternative setup where the method is provided by an additional driving image with similar pose and expression as the source. This additional driving image was utilised to obtain better approximation of the identity independent motion of the driving face. Unfortunately, such a driving image would be challenging to obtain.

\vspace{-0.5cm}
\paragraph{Deep Generative Adversarial Networks}
The Generative Adversarial Networks (GANs) are popular models for high quality image generation \cite{karrasStyleBasedGeneratorArchitecture2019,karrasProgressiveGrowingGANs2018} and manipulation \cite{bauGANDISSECTIONVISUALIZING2019d,wangHighResolutionImageSynthesis2018}. In \cite{zakharovFewShotAdversarialLearning2019,nirkinFSGANSubjectAgnostic2019}, a progressively growing GAN architecture was trained to animate a source face based on the landmarks of the driving face. As landmarks preserve the facial structure and, up to some extent, the identity, these models suffer from quality degradation if the driving and source faces are not similar. However, this approach is well suited for applications like telepresence where the driving and source identities are the same.

Landmark transfer models were proposed in \cite{wuReenactGANLearningReenact2018} and \cite{haMarioNETteFewshotFace2019} to remove the identity features from the driving landmarks. In \cite{wuReenactGANLearningReenact2018}, the landmark transformer was trained separately for each identity pair, which limited the scalability of the method. A principal component analysis (PCA) based model was used in \cite{haMarioNETteFewshotFace2019} to separate expression and shape parameters from the landmarks. They utilised shape parameters from the source and expression parameters from the driving face to reconstruct the transferred landmarks. Although the method generalizes better compared to \cite{wuReenactGANLearningReenact2018}, the proposed linear models are not sufficient to differentiate complex expressions from the shape information.

An action units (AUs) based face representation is used in \cite{pumarolaGANimationAnatomicallyawareFacial2018} to manipulate facial expressions (not pose). More recently, in \cite{tripathyICfaceInterpretableControllable2020}, the authors proposed a model that used AUs for the full face reenactment (expression and pose). The AUs represent complex facial expressions by modeling the specific muscle activities \cite{ekmanFacialActionCoding1978}. These activations are independent of the facial structure, which makes the action units a potential representation to disentangle driving motion from the identity. Unfortunately, the model presented in \cite{tripathyICfaceInterpretableControllable2020} was not able to produce similar output quality as the facial landmark based alternatives.

In this paper, we combine the disentanglement properties of the action units with strong supervision from the facial landmarks to produce a high quality reenactment results without significant identity leakage problems.

\section{Method}
\label{sec:orgd635f9e}
\begin{figure*}[!t]
\begin{center}
     \includegraphics[width=0.95\linewidth]{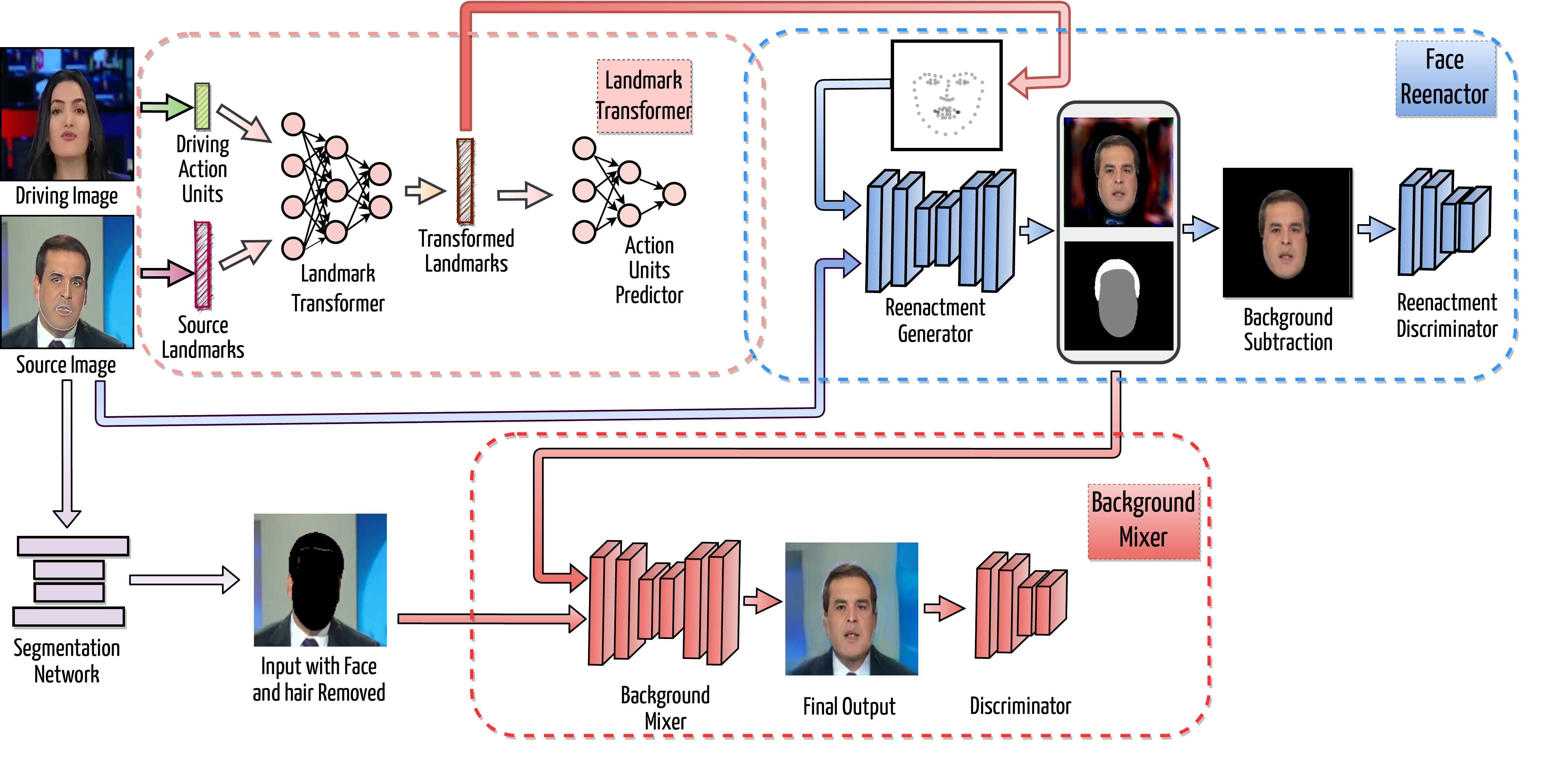}
\end{center}
   \caption{The overall architecture of the proposed FACEGAN contains three major blocks, namely 1. Landmark transformer, 2. Face reenactor, and 3. Background Mixer. The Landmark transformer modifies the source landmarks to include the driving face's expression and pose, which are represented in terms of action units (AUs). The modified landmarks and the source image are provided to the Face reenactor module for producing the reenacted face. Finally, the reenacted face and the source background are combined into the output by the Background mixer module. In the training phase, we use the source and driving images from the same face track, which allows us to use the driving image as a pixel-wise ground truth of the output. However, during the inference, the source and driving may have different identities.}
\label{reenact_block_dig}
\vspace{-0.3cm}
\end{figure*}

Given an image, \(I_d\) with a driving face \(F_d\) and an image \(I_s\) with a source face \(F_s\), the FACEGAN model aims to transfer the motion (pose and expression) from \(F_d\) to \(F_s\) while maintaining the identity of \(F_s\). To do so, the model takes the landmarks of \(F_s\) and manipulates them to include the motion of \(F_d\). The motion information is extracted from \(F_d\) in terms of action units (AUs), which correspond to various face muscle activations \cite{ekmanWhatFaceReveals1997a} and head pose angles. The overall shape features of the source landmarks are not altered during the transformation, which helps to preserve the source identity. The transformed landmarks are passed to reenactment and background mapper modules, which together generate a face image depicting the source identity and background, but with driving facial motion.

The overall architecture of FACEGAN contains three main components, which are illustrated in figure \ref{reenact_block_dig}. The first component, called landmark-transformer, consisting of a fully connected network \(L_T\). It takes the landmarks \(l_s\) from \(F_s\) and the action units \(AU_d\) of \(F_d\) as inputs and generates the transformed landmarks \(L_t\) as an output. \(L_t\) has pose and expression from \(F_d\), whereas the facial structure from \(F_s\) (as the input landmarks are from \(F_s\)).

The transformed landmarks \(L_t\) are converted to a gray scale image \(H_t\), as shown in Figure \ref{reenact_block_dig}, and provided to the next component called as face-reenactor. The face-reenactor takes \(I_s\) and \(H_t\) as inputs for the generator \(G_r\) and produces the reenacted face \(I_{fr}\) as an output. \(G_r\) contains also a segmentation unit \(G_{rs}\), which provides a face, hair and background segmentation maps of \(I_{fr}\).

By removing the background from \(I_{fr}\) and the face from \(I_s\), we obtain two images called \(I_f\) and \(I_b\). These images are provided to the third module called background mixer. \(I_f\) contains the reenacted face information whereas \(I_b\) contains the source background. Unlike other contemporary approaches, our model processes the background separately from the face area. In this way, different components will specialize to face reenactment and background manipulation, leading to improved output quality. The background mixer contains a generator \(G_b\) producing \(I_r\), which has facial identity and background from \(I_s\) but facial pose and expression from \(I_d\).

The training process of our model requires image frames extracted from talking head videos. In the following, we assume that the facial landmarks of the source faces, and the head poses and the action units (AUs) of the driving faces, are already extracted and stored. This can be achieved using publicly available landmark \cite{bulatHowFarAre2017} and Pose-AU detectors \cite{baltrusaitisOpenfaceFacialBehavior2018,baltrusaitisCrossdatasetLearningPersonspecific2015}. For simplicity, we denote the AU and pose combination as AUs throughout the text unless mentioned otherwise. In each step of the training procedure, we extract two frames \(I_s\) and \(I_d\) from the same video track as described in \cite{wilesX2FaceNetworkControlling2018,tripathyICfaceInterpretableControllable2020,zakharovFewShotAdversarialLearning2019}. In this way, the exact ground truth of our reenacted source is available in terms of I\textsubscript{d}. The detailed training steps and loss functions are discussed in the following sections.

\subsection{Landmark Transformer}
\label{sec:orge05cf27}

The landmark transformer \(L_T\) modifies the source landmarks according to the driving motion, while maintaining the facial shape and identity of the source face. First, \emph{2D} facial landmarks are extracted from the source image \(I_s\) \(\in\) \(\mathbf{R}^{3 \times H \times W}\) and these are subsequently reshaped to \(l_s\) \(\in\) \(\mathbf{R}^{136 \times 1}\). The facial motion of \(I_d\) \(\in\) \(\mathbf{R}^{3 \times H \times W}\) is extracted in terms of the action units (\(\in\) \(\mathbf{R}^{17 \times 1}\)) and the head pose angles (\(\in\) \(\mathbf{R}^{3 \times 1}\)). These two parameters are concatenated in to a single vector \(AU_d\) \(\in\) \(\mathbf{R}^{20 \times 1}\) that describes the complete motion of \(F_d\). The landmark transformer \(L_T\) takes the concatenated \(l_s\) and \(AU_d\) as an input and predicts the landmark movements \(\delta l_s\). Finally, the transformed source landmarks are obtained as \(l_t = l_s + \delta l_s\).

The landmark transformer is trained using various loss functions. The most straightforward loss is calculated between \(l_t\) and the landmarks of \(I_d\) denoted as \(l_d\). Since during training, \(I_s\) and \(I_d\) are from the same video track, the landmarks \(l_d\) form the ground truth for \(l_t\). To smooth the predictions \(\delta l_s\) a l2-weighted penalty regularization is added along with the reconstruction loss. The objective function can be written as,
\begin{equation}
 \mathcal{L}_{lr} =\vert \vert l_t-l_d \vert \vert_1 + \lambda_{lr} \vert \vert \delta _s \vert \vert_2.
 \label{equ:landmark_reconstruction}
\end{equation}

Although \(\mathcal{L}_{lr}\) encourage the proper pose variations in \(l_t\), they fail to capture the subtle movements due to the expression variations. In order to focus on expression, another fully connected network \(L_a\) is introduced to regress the AU parameters from the landmarks. \(L_a\) is trained simultaneously with \(l_t\) using the following loss function
\begin{equation}
 \mathcal{L}_{lau} =\vert \vert L_a(l_t)-AU_d \vert \vert_1 + \vert \vert L_a(l_d)-AU_d \vert \vert_1.
 \label{equ:landmark_AU_reconstruction}
\end{equation}

In order to preserve the facial shape in landmark domain, we include a connectivity loss. The loss preserves the distances between the neighbouring landmark points in \(l_s\) and \(l_t\).  The loss function is mathematically expressed as,
\begin{equation}
 \mathcal{L}_{lc} =\vert \vert D_t-D_d \vert \vert_1,
 \label{equ:landmark_connectivity}
\end{equation}
where \(D_t\) \(\in\) \(\mathbf{R}^{d \times 1}\) and \(D_d\) \(\in\) \(\mathbf{R}^{d \times 1}\) are vectors containing the differences between the connected landmark positions of $l_t$ and $l_d$. Here \(d\) is the number of connected landmarks included in the loss function calculation. Finally, the full loss function for the landmark transformer is formed by a linear combination of equations (\ref{equ:landmark_reconstruction}), (\ref{equ:landmark_AU_reconstruction}), and (\ref{equ:landmark_connectivity}) as
\begin{equation}
 \mathcal{L}_{l} = \lambda_{l1} \mathcal{L}_{lr} + \lambda_{l2} \mathcal{L}_{lau} + \lambda_{l3} \mathcal{L}_{lc},
 \label{equ:landmark_complete}
\end{equation}
where \(\lambda_{l1}\), \(\lambda_{l2}\), and \(\lambda_{l3}\) are weighting constants.

\subsection{Face Reenactor}
\label{sec:org5a97333}

The transformed landmarks \(l_t\) are mapped to single channel image \(H_t\) \(\in\) \(\mathbf{R}^{1 \times H \times W}\) by placing a 2D Gaussian function at each keypoint location as shown in figure \ref{reenact_block_dig}. \(H_t\) is channel-wise concatenated with the source image \(I_s\) to form the input for the face reenactor network \(G_r\). \(G_r\) produces a reenacted RGB image \(I_{fr}\) \(\in\) \(\mathbf{R}^{3 \times H \times W}\) and a segmentation map \(S_{fr}\) \(\mathbf{R}^{3 \times H \times W}\) with face, hair and background classes. The segmentation maps are obtained using a CNN head added to the second last layer of the \(G_r\). For training, the pixel wise reconstruction loss, VGG perceptual loss \cite{wangHighResolutionImageSynthesis2018}, and adversarial loss are applied as given in equations (\ref{equ:reenactment_pixel}), (\ref{equ:reenactment_percep}), and (\ref{equ:reenactment_adv}). The segmentation branch is trained with standard cross entropy loss. The full loss function for \(G_r\) is provided in Equation (\ref{equ:reenactment+full}).

\begin{equation}
 \mathcal{L}_{rr} =\vert \vert I_{fr}^{'} - F_d \vert \vert_1
 \label{equ:reenactment_pixel}
\end{equation}
\begin{equation}
 \mathcal{L}_{rp} = \sum_i \frac{1}{C_iH_iW_i} \vert \vert V_i(I_{fr}^{'}) - V_i(F_d) \vert \vert_1
 \label{equ:reenactment_percep}
\end{equation}
\begin{equation}
\begin{aligned}
 \mathcal{L}_{radv} ={} & \min_{G_r} \max_{D_{r1}, D_{r2}, D_{r3}} \sum_{r=1}^{3} \mathbb{E}_{F_d}\left[ log D_r(F_d)\right] \\
		       & + \mathbb{E}_{L_S}\left[ log( 1-D_r(I_{fr}^{'}))\right]
 \label{equ:reenactment_adv}
\end{aligned}
\end{equation}
\begin{equation}
 \mathcal{L}_{r} = \lambda_{r1} \mathcal{L}_{rr} + \lambda_{r2} \mathcal{L}_{rp} + \lambda_{r3} \mathcal{L}_{radv} + \lambda_{r4} \mathcal{L}_{ce}.
 \label{equ:reenactment+full}
\end{equation}

\(I_{fr}^{'}\) and \(F_d\) are \(I_{fr}\) and \(I_d\) with background removed. Each \(D_r\) in equation (\ref{equ:reenactment_adv}) stands for discriminator, used for multiple resolution of images as described in \cite{wangHighResolutionImageSynthesis2018}. \(\mathcal{L}_{ce}\) is a standard cross entropy loss for \(S_{fr}\). In order to obtain the ground truth for \(\mathcal{L}_{ce}\), a pretrained face-segmentation network \(G_s\) is utilized as explained in \cite{nirkinFaceSegmentationFace2017}.

\subsection{Background Mixer}
\label{sec:org466ac72}

The input to the background mixer network \(G_b\) is a channel-wise
concatenation of \(I_{fr}^{'}\) and \(I_b\), where \(I_b\) is \(I_s\)
with face and hair removed. \(G_b\) generates an RGB image \(I_c\) and
a single channel mask \(M\) \(\in\) \(\mathbf{R}^{HXW}\). Finally, the reenacted output image \(I_r\) is obtained using \(I_c\), \(I_b\), and \(M\) as

\begin{equation}
 I_r = M * I_c + (1-M) * I_b.
 \label{equ:background_mix}
\end{equation}

In this way, \(G_b\) is encouraged to focus on producing the compatible background while directly copying as much information form \(I_b\) as possible. \(G_b\) is trained using pixel loss and adversarial loss on \(I_r\), and an additional regularization for \(M\). The regularization is obtained by imposing a \(l_2\) weight penalty, smoothing the mixing process in equation \ref{equ:background_mix}, and applying a total variation regularization. The full loss on \(M\) can be written as,

\begin{equation}
  \begin{aligned}
   \mathcal{L}_{bm} ={} & \lambda_{b1} \sum_{x,y}^{H,W} [(M^{x+1, y} - M^{x,y})^2 + (M^{x, y+1} - M^{x,y})^2] \\
                       & + \lambda_{b2} \vert \vert M \vert \vert_2
   \label{equ:background_mix}
  \end{aligned}
\end{equation}

Finally, the complete loss function for the background transformer with the regularization parameters can be written as,
\begin{equation}
 \mathcal{L}_{b} = \mathcal{L}_{bm} + \lambda_{b3} \mathcal{L}_{bp} + \lambda_{b4} \mathcal{L}_{badv} + \lambda_{b5} \mathcal{L}_{br},
 \label{equ:background_full}
\end{equation}
where \(\mathcal{L}_{bp}, \mathcal{L}_{br}\) are perceptual loss and reconstruction loss between \(I_r\) and \(I_d\), respectively. \(\mathcal{L}_{badv}\) is the adversarial loss of the Equation (\ref{equ:reenactment_adv}) for the discriminator \(D_b\) and \(G_b\).

\section{Training Details}
\label{sec:org49933af}

We train FACEGAN using a dataset created from IJB-C videos \cite{mazeIARPAJanusBenchmark2018}. Most of these videos contain celebrities talking in an unconstrained setup. The faces are first detected using a landmark detector and then tracked using a centroid tracker over the video frames. Around the centroids, a fixed image crop is used to extract the faces in a way that the middle position between the eyes will always be in a fixed position. Small faces are rejected based on landmarks height and width. In total 400k good quality face images are obtained. In addition, we used 400 videos from the Forensic++ dataset \cite{rosslerFaceForensicsLearningDetect2019} to evaluate the performance of our system. Forensic++ was pre-processed in same way as IJB-C, resulting in total \(200k\) face images.

The networks \(G_r\) and \(G_b\) have U-Net like structures and are trained progressively as given in \cite{wangHighResolutionImageSynthesis2018}. We first trained the networks to generate images of resolution \(128 \times 128\) and then increased to the final resolution of \(256 \times 256\). The landmark transformer \(L_T\) is a fully connected network predicting landmark locations in normalized coordinates. These can be converted to a single channel image of any resolution. All networks are trained separately for a few epochs and then together in and end-to-end manner. For \(L_T\) and for other generator networks, the learning rate is kept at 0.0002 whereas the batch-size is 32 for the former and 1 for the latter. %

\section{Experiments}
\label{sec:org80f989f}

We assess the proposed approach in a series of experiments and compare the results against the recent state-of-the-art works \cite{wilesX2FaceNetworkControlling2018,nirkinFSGANSubjectAgnostic2019,siarohinFirstOrderMotion2019,tripathyICfaceInterpretableControllable2020} in quantitative and qualitative terms. We start by evaluating our landmark transformer in isolation and then continue with the full face reenactment comparisons. All experiments are performed using the \(200k\) face images obtained from the FaceForensic++ dataset \cite{rosslerFaceForensicsLearningDetect2019} as described in Section \ref{sec:org49933af}. We also note that none of the tested models, including ours, are trained with FaceForensic++, which emphasizes the generalization properties of the models. In all cases, we use one source image per identity and the final output resolution is \(256 \times 256\).

We compare our approach with four recent works, namely X2face \cite{wilesX2FaceNetworkControlling2018}, FSGAN \cite{nirkinFSGANSubjectAgnostic2019}, First-Order-motion-Model(FOM) \cite{siarohinFirstOrderMotion2019} and ICface \cite{tripathyICfaceInterpretableControllable2020}. These state-of-the-art models are representatives from a broader set of reenactment frameworks. For instance, FSGAN and ICface \cite{tripathyICfaceInterpretableControllable2020} utilise similar ideas of interpretable facial attributes like landmarks and AUs in reenactment. In contrast, X2face \cite{wilesX2FaceNetworkControlling2018} and First-order-motion-model \cite{siarohinFirstOrderMotion2019} represent works that learn motion and identity representation in an unsupervised (or self-supervised) setups. By comparing FACEGAN with these representatives from both categories, we obtain better understanding of the drawback in the existing works and the contributions of FACEGAN. We used the original source codes for all comparison methods. We did not consider the few-shot learning models \cite{zakharovFewShotAdversarialLearning2019} and \cite{haMarioNETteFewshotFace2019} since they require inference time training and the corresponding implementations are not available.

\begin{figure}[!t]
\begin{center}
     \includegraphics[width=1.05\linewidth]{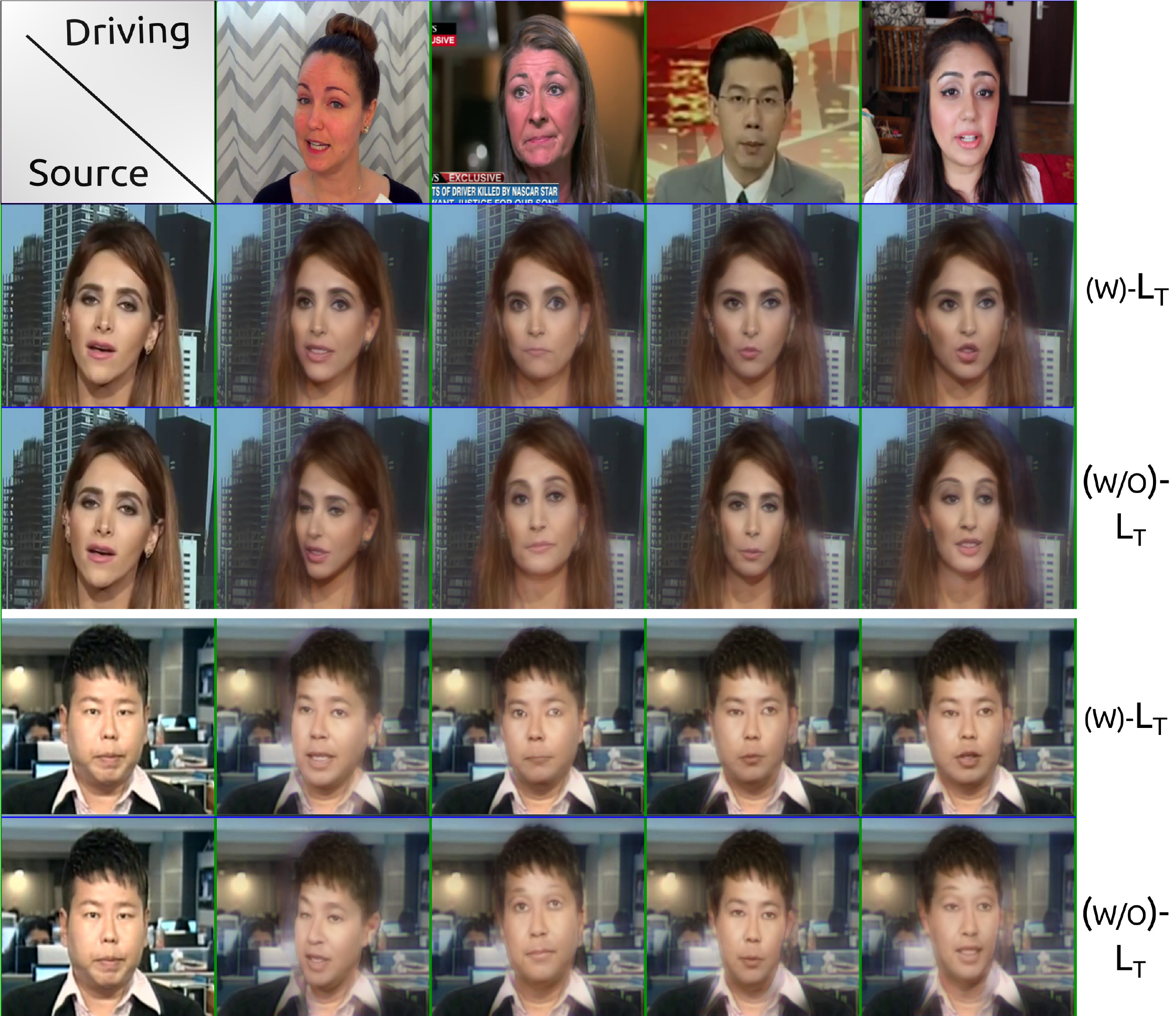}
   \caption{\label{fig:land_no_land} Qualitative comparison of FACEGAN outputs with and without the landmark transformer. The facial shape and identity are clearly distorted if the landmark transformer is not applied.}
\end{center}
\vspace{-0.5cm}
\end{figure}

\begin{figure*}[!t]
\begin{center}
     \includegraphics[width=0.9\linewidth]{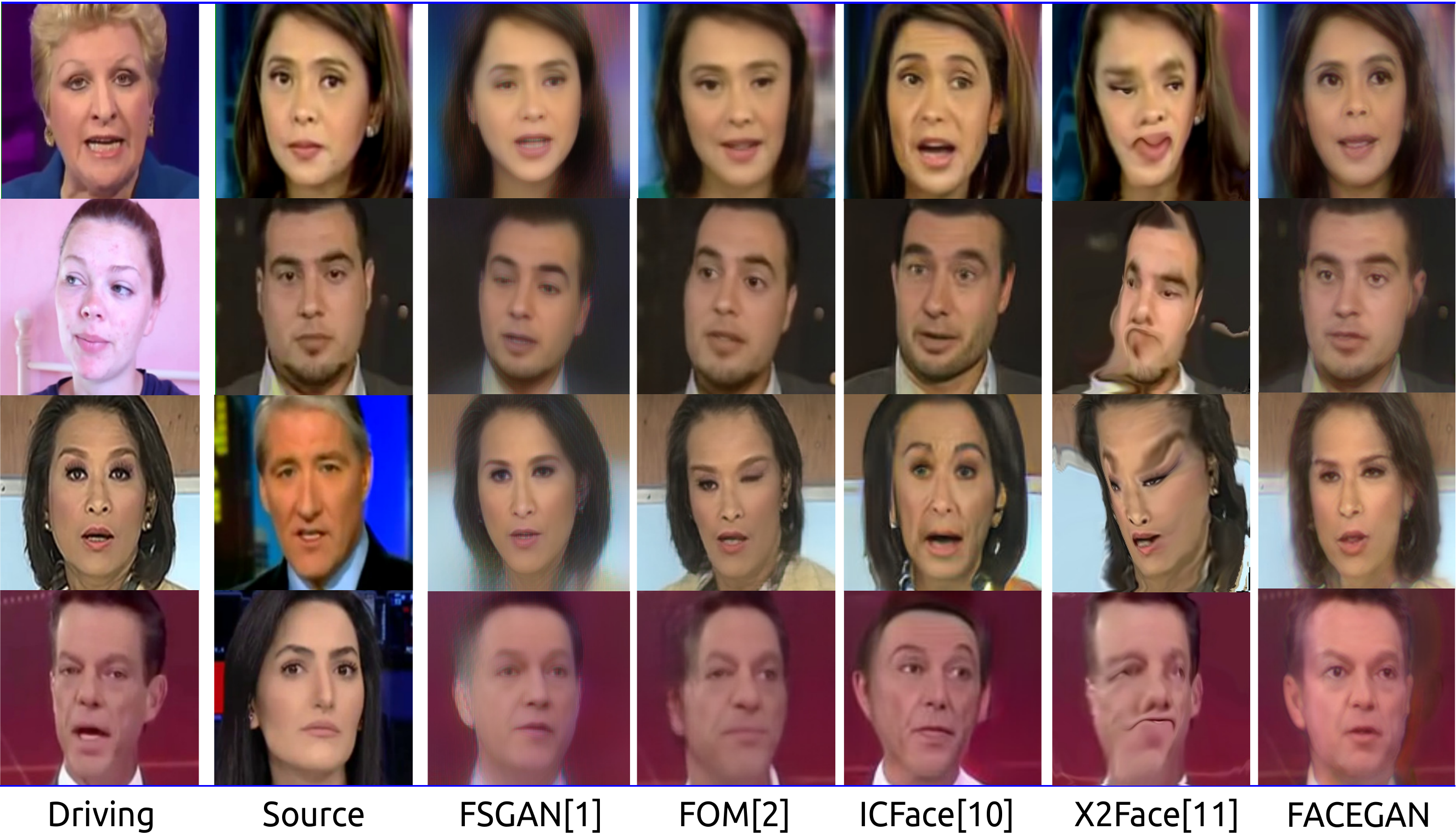}

   \caption{\label{fig:rcomp} Qualitative comparison of FACEGAN and four recent baseline methods: FSGAN \cite{nirkinFSGANSubjectAgnostic2019}, First-Order-motion-Model(FOM) \cite{siarohinFirstOrderMotion2019}, ICface \cite{tripathyICfaceInterpretableControllable2020}, and X2face \cite{wilesX2FaceNetworkControlling2018}. FACEGAN is clearly able to retain the source face shape and identity better compared to the baselines. Moreover, the driving motion is faithfully reproduced.}
\end{center}
\vspace{-0.8cm}
\end{figure*}

\subsection{Landmark Transformer}
\label{sec:orgb51b1ba}

In order to evaluate the importance of the landmark transformer (\(L_T\)) in our model, we performed self-reenactment and cross-reenactment with and without \(L_T\). The self-reenactment refers to a case where the source and driving images are from the same identity whereas in cross-reenactment these identities are different. In the former case, we take the source and driving images from the same face track, which allows us to use the driving image as a pixel-wise ground truth of the reenacted output. Here we evaluate the result using a pixel-wise with Mean Square euclidean Loss (MSE). In the cross-reenactment case, we do not have access to the ground truth image and cannot use the MSE loss. Instead, inspired by \cite{haMarioNETteFewshotFace2019,zakharovFewShotAdversarialLearning2019,xiangOneShotIdentityPreservingPortrait2020}, we calculate the following three measures

1. \emph{Cosine Similarity between IMage embeddings (CSIM):} uses a pre-trained face recognition network \cite{parkhiDeepFaceRecognition2015} to obtain embeddings from the source and driving images. Higher score indicates that the source identity is better preserved in the reenactment process.
2. \emph{Pose Cosine Similarity between IMage (PSIM):} is calculated between the head pose angles, estimated using Openface \cite{baltrusaitisOpenfaceFacialBehavior2018}, of the driving and the reenacted faces. PSIM measures the ability of the model to retain the driving head poses in the final output.
3. \emph{Expression difference (ED):} is the Euclidean distance between the action units, calculated with Openface \cite{baltrusaitisOpenfaceFacialBehavior2018}, of driving and reenacted images. It measures the ability of the model to retain the driving expressions in the final image.

The results of the landmark transformer experiments are presented in Table \ref{tabel-landmarks}. In the self-reenactment case, the best performance is obtained without using the landmark transformer. The result is expected, since in the self-reenactment setup, the driving landmarks represent the optimal output of the landmark transformer. Nevertheless, the MSE score is only slightly lower if the landmark transformer is applied, which indicates that the expression and pose are faithfully transferred from the driving face.

In the cross-reenactment case, we observe a significant difference in the CSIM score while the PSIM and ED scores are similar. High CSIM score indicates that the source identity is well preserved in the reenactment output. The large difference in terms of CSIM shows that the landmark transformer clearly improves the reenactment quality and reduces the identity leakage problem. The PSIM and ED scores measure pose and expression similarity between the output and the driving face. These measures do not depend on the face identity and the driving landmarks directly provide strong supervision in these terms. However, the landmark transformer results in similar or better scores also according to these measures, which indicates that the pose and expression are faithfully transferred despite substantial improvement in preserving the source identity. Figure \ref{fig:land_no_land} shows a few examples of the cross-reenactment results for FACEGAN with and without the landmark transformer. These illustrate in particular how the face structure is preserved better by using the landmark transformer.

\begin{table}[!t]
\centering
\begin{tabular}{p{2.75cm} p{1.7cm} | p{0.6cm} p{0.6cm} p{0.7cm}}
\hline
 & Self-Reenactment & Cross-Reenactment   & &\\
\hline
Model & MSE$\downarrow$ & CSIM$\uparrow$ & PSIM$\uparrow$ & ED$\downarrow$\\
\hline
FACEGAN w $L_T$ & 0.018 & \textbf{0.747} & 0.811 & \textbf{0.012}\\
\hline
FACEGAN w/o $L_T$ & \textbf{0.015} & 0.623 & \textbf{0.885} & 0.014\\
\hline
\end{tabular}
\caption{Quantitative evaluation of the impact of the Landmark transformer($L_T$). High CSIM and PSIM scores indicate the ability to preserve the source identity and driving head pose, respectively. Low ED score reflects the ability to reproduce the driving expression. The MSE score measures the pixel-wise differences in the self-reenactment case.}
\label{tabel-landmarks}
\vspace{-0.6cm}
\end{table}

\subsection{Qualitative Reenactment Results}
\label{sec:orgf61c6ae}

In this section, we assess the reenactment performance of the proposed model in qualitative terms and compare it with X2face \cite{wilesX2FaceNetworkControlling2018}, FSGAN \cite{nirkinFSGANSubjectAgnostic2019}, First-Order-motion-Model(FOM) \cite{siarohinFirstOrderMotion2019} and ICface \cite{tripathyICfaceInterpretableControllable2020}. The results for versatile driving-source pairs are illustrated in figures \ref{fig:rcomp} and \ref{fig:org1574033}. The source identity is well preserved in the FACEGAN results, and there is considerably less visible structure leakage compared to the baseline methods. The difference is particularly due to the landmark transformer, which helps to preserve the shape of the source face.

In addition, most models like X2Face \cite{wilesX2FaceNetworkControlling2018}, FSGAN \cite{nirkinFSGANSubjectAgnostic2019}, and ICFace \cite{tripathyICfaceInterpretableControllable2020} operate only on a tight crop around the face area. Such approach completely ignores the untrivial integration of face, background, and other body parts such as hair. These limitations greatly hamper the practical usability of the methods. In contrast, FACEGAN has a dedicated background mixer model that generates the context for the reenacted face by hallucinating high quality background pixels along with the ears and the upper body parts as shown in figures \ref{fig:land_no_land} and \ref{fig:orge55ddfd} (Note that the background is cropped out from FACEGAN results in figures \ref{fig:rcomp} and \ref{fig:org1574033} to facilitate comparison). The background mixer enables the reenactment network to focus on the face and hair regions improving the reenactment quality of these parts. In conclusion, our model produces sharper and better reenactment results from a single source and driving images of different identity in comparison to the state-of-the-art models.

\subsection{Quantitative Reenactment Results}
\label{sec:org3d645f2}

In this section, we compare the FACEGAN results with the baseline models in quantitative terms. We use similar self and cross-reenactment setups as in Section \ref{sec:orgb51b1ba} and report the results using MSE, CSIM, PSIM and ED scores in Table \ref{table-comparision}. In the self-reenactment case, all methods except ICFace obtain similar MSE scores. Comparison to ICFace demonstrates the benefits of using the facial landmark representation instead of pure action units. In the cross-reenactment setup, FACEGAN obtains the highest performance in terms of the CSIM score with a clear margin. This result further illustrates the ability of FACEGAN to preserve the source identity faithfully. In terms of PSIM, the direct landmark based FSGAN model results in the best pose retention capability. However, FACEGAN and FOM are not far behind the FSGAN. ICFace has the lowest PSIM score, which further illustrates the difficulty of producing faithful pose using purely action unit based supervision. In terms of ED metrics, FACEGAN obtains the best results, followed by ICFace.

While the CSIM score reflects the overall ability to preserve the identity, it may be inadequate to reflect small scale structural differences. The difficulty, however, would be to obtain a sufficient ground truth for measuring such details. For this purpose, we propose a new measure called landmark similarity score (LSIM). To calculate the measure, we first randomly choose two source images from a single identity. Then we search the test database for a driving image with different identity but as similar action units as possible to the second source image. Finally, we use the motion information (AUs, landmarks, warping parameters) from the discovered driving image to reenact the face in the first source image. The output is compared with the second source image that acts as a pseudo ground truth for the output. To accommodate for a small differences, we calculate LSIM as a mean square error between the corresponding landmark locations instead of pixel level MSE. If the reenactment process preserves the facial shape and motions accurately then it should result in low LSIM score. Table \ref{table-comparision} contains the obtained results. The proposed FACEGAN model achieves clearly the highest performance among the compared approaches.

\begin{table}[htbp]
\centering
\begin{tabular}{p{1.75cm} p{1.7cm} | p{0.6cm} p{0.6cm} p{0.6cm} p{0.5cm}}
\hline
  & Self-Reenactment & & Cross-Reenactment   & &\\
Model & MSE$\downarrow$ & CSIM$\uparrow$ & PSIM$\uparrow$ & ED$\downarrow$ & LSIM$\downarrow$ \\
\hline
X2Face \cite{wilesX2FaceNetworkControlling2018} & 0.018 & 0.564 & 0.659 & 0.025 & 0.040 \\
\hline
FSGAN \cite{nirkinFSGANSubjectAgnostic2019} & 0.016 & 0.631 & \textbf{0.887} & 0.016 & 0.035 \\
\hline
ICFace \cite{tripathyICfaceInterpretableControllable2020} & 0.106 & 0.655 & 0.699 & 0.013 & 0.039\\
\hline
FOM \cite{siarohinFirstOrderMotion2019} & \textbf{0.012} & 0.676 & 0.825 & 0.015 & 0.035\\
\hline
FACEGAN  & 0.018 & \textbf{0.747} & 0.811 & \textbf{0.012} & \textbf{0.021}\\
\hline
\end{tabular}
\caption{Quantitative comparison of our model with the state-of-the-arts works. High CSIM scores indicate better ability to preserve the source identity, while low LSIM signifies better landmark shape retention ability. The PSIM and ED measure the head pose and expression reproduction, respectively. FACEGAN obtains clearly the best CSIM and LSIM scores, while obtaining similar PSIM and ED scores. This indicates that our method is capable of preserving the source identity while faithfully reproducing the driving motion. }
\label{table-comparision}
\end{table}

\subsection{Controllable face reenactment}
\label{sec:org9169fc4}

\begin{figure}[!t]
\centering
\includegraphics[width=0.5\textwidth]{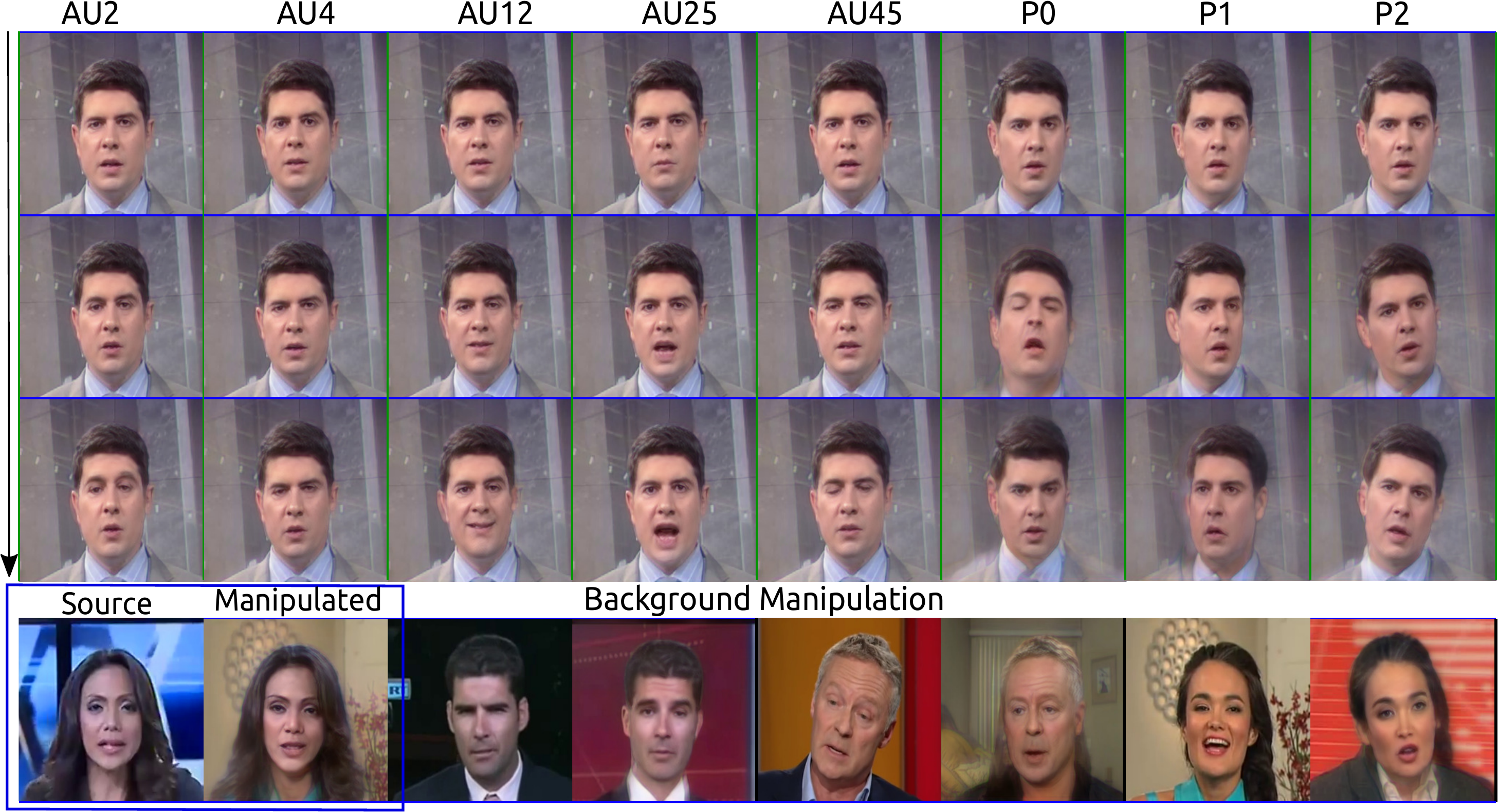}
\caption{\label{fig:orge55ddfd}Demonstration of the selective editing capabilities of FACEGAN. The first three rows images are generated by increasing the source AUs directly from minimum to maximum value. The last column illustrate the capability of combining the reenacted face with a non-source background (zoom in to observe the quality better).}
\vspace{-0.5cm}
\end{figure}

FACEGAN utilises \(20\) human interpretable action units (AUs) to represent the pose and expression of the desired output face. The AUs can be obtained from a driving face, but this is not the only option. For instance, one can take the AUs from the source face, manipulate them, and feed them as driving information to the model. In this way, one can selectively edit the source image by, for example, changing the pose or adding an extra smile. Figure \ref{fig:orge55ddfd} contains a few examples of such selective editing procudure.

Moreover, the background mixer can combine face and background from two different sources. This results in a mixed reenactment result, where the face identity is from one source and the background from another. Figure \ref{fig:orge55ddfd} illustrates a few examples of this kind of edits. Such feature would be very useful for relocating the source person into a desired environment. The proposed selective editing properties provide complete freedom and control to generate the desired reenactment video.

\section{Conclusion}
\label{sec:org8eed2ff}

We proposed a facial animator called FACEGAN that is capable of performing high quality reenactment from a single source image. Unlike many previous works, our model does not pose any restriction on the compatibility of the source and driving pairs. The model combines the best properties of the action unit and facial landmark motion representations for reducing the identity leakage problem and to optimise the reenactment quality. Furthermore, FACEGAN handles the face and the background separately which improves the output quality and gives additional control of choosing the desired background. We have compared our method with the state-of-the-art approaches and obtained superior results both quantitatively and qualitatively.

\bibliographystyle{splncs}
\bibliography{bmvc2020}
\end{document}